\documentclass{article}

\usepackage{arxiv}

\usepackage[utf8]{inputenc} 
\usepackage[T1]{fontenc}    
\usepackage{hyperref}       
\usepackage{url}            
\usepackage{booktabs}       
\usepackage{amsfonts}       
\usepackage{nicefrac}       
\usepackage{microtype}      
\usepackage{lipsum}		
\usepackage{graphicx}
\usepackage{amsmath}
\usepackage{amssymb}
\usepackage{subcaption}
\usepackage{caption}
\makeatletter
\newcommand{\clearsubcaptcounter}{\setcounter{sub\@captype}{0}}
\makeatother

\title{Foreground-guided Facial Inpainting with Fidelity Preservation}

\author{ \href{https://orcid.org/0000-0003-2309-4655}{\includegraphics[scale=0.06]{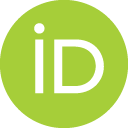}\hspace{1mm}Jireh Jam $^{1}$}\\
	\And
	\href{https://orcid.org/0000-0002-3623-6598}{\includegraphics[scale=0.06]{orcid.png}\hspace{1mm}Connah Kendrick $^{1}$}\\
	\And
	\href{https://orcid.org/0000-0001-7055-9609}{\includegraphics[scale=0.06]{orcid.png}\hspace{1mm}Vincent Drouard $^{2}$} \\
	\And
	\href{https://orcid.org/0000-0002-3009-3311}{\includegraphics[scale=0.06]{orcid.png}\hspace{1mm}Kevin Walker $^{2}$} \\
	\And
	\href{https://orcid.org/0000-0001-7681-4287}{\includegraphics[scale=0.06]{orcid.png}\hspace{1mm}Moi Hoon Yap $^{1}$}\thanks{The authors would like to thank The Royal Society (Grant number: IF160006 and INF/PHD/180007). We gratefully acknowledge the support of NVIDIA Corporation with the donation of the Quadro P6000 used for this research.} \\
	\And
	$^{1}$ Manchester Metropolitan University, \\
	Manchester, United Kingdom\\
	\And
	$^{2}$ Image Metrics Ltd,\\
	Manchester, United Kingdom \\
}

\date{}

\renewcommand{\headeright}{}
\renewcommand{\undertitle}{}

\begin{document}
\maketitle

\begin{abstract}
	Facial image inpainting, with high-fidelity preservation for image realism, is a very challenging task. This is due to the subtle texture in key facial features (component) that are not easily transferable. 
	Many image inpainting techniques have been proposed with outstanding capabilities and high quantitative performances recorded. However, with facial inpainting, the features are more conspicuous and the visual quality of the blended inpainted regions are more important qualitatively. Based on these facts, we design a foreground-guided facial inpainting framework that can extract and generate facial features using convolutional neural network layers. It introduces the use of foreground segmentation masks to preserve the fidelity. Specifically, we propose a new loss function with semantic capability reasoning of facial expressions, natural and unnatural features (make-up). We conduct our experiments using the CelebA-HQ dataset, segmentation masks from CelebAMask-HQ (for foreground guidance) and Quick Draw Mask (for missing regions). Our proposed method achieved comparable quantitative results when compare to the state of the art but qualitatively, it demonstrated high-fidelity preservation of facial components.
\end{abstract}

\begin{figure}[!ht]
\centering
\begin{subfigure}[b]{0.16\linewidth}        
	\centering
	\includegraphics[width=\linewidth]{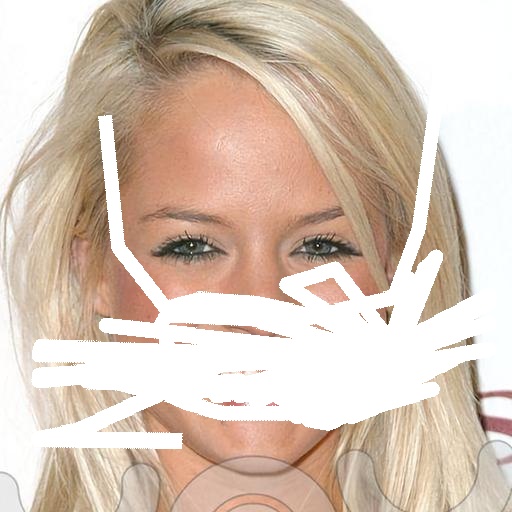}
	\caption{Input}
\end{subfigure}
\begin{subfigure}[b]{0.16\linewidth}        
	\centering
	\includegraphics[width=\linewidth]{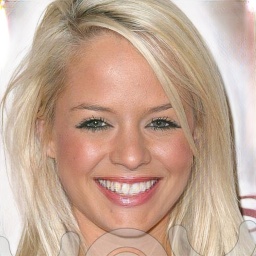}
	\caption{Ours}
\end{subfigure}
\begin{subfigure}[b]{0.16\linewidth}        
	\centering
	\includegraphics[width=\linewidth]{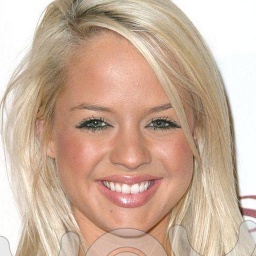}
	\caption{GT}
\end{subfigure}
\clearsubcaptcounter	
\begin{subfigure}[b]{0.16\linewidth}        
	\centering
	\includegraphics[width=\linewidth]{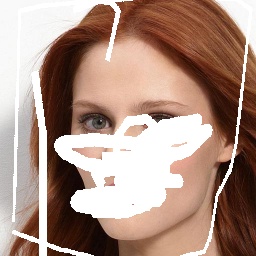}
	\caption{Input}
\end{subfigure}
\begin{subfigure}[b]{0.16\linewidth}        
	\centering
	\includegraphics[width=\linewidth]{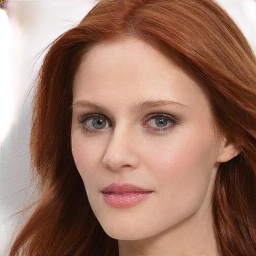}
	\caption{ours}
\end{subfigure}
\begin{subfigure}[b]{0.16\linewidth}        
	\centering
	\includegraphics[width=\linewidth]{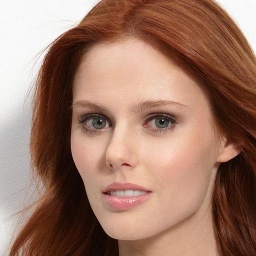}
	\caption{GT}
\end{subfigure}
\caption{\label{fig:ours} Inpainted images (b) from our model showing semantic understanding with preserved realism. The predictions of targeted regions show our model can image preserve realism.}
\end{figure}
\section{Introduction}
Image inpainting is an ongoing challenging research in computer vision. It is the art of reconstructing an image using algorithms powered by complex mathematical computations running in the background. Image inpainting is widely known as a restoration or editing technique commonly used by image/video editing applications. Generative neural networks have shown greater ability to extract features over conventional (traditional) methods. The later usually use low level features to fill-in missing pixels by diffusion or match by exemplars to complete an inpainting task. The former use CNNs, capable of extracting high-quality textural and structural information that can fill in the missing contents by training a large scale dataset in a data-driven manner. The CNN model is known in literature to predict and understand an image structure without an explicit modelling of structures during the learning process \cite{xiong2019foreground}. 

Inpainting is not an easy task for ongoing learning-based models and have several limitations, some of which have been mentioned in Jam et al. \cite{jam2020comprehensive}. One reason is the difficulty to propagate features from one area of the feature map to another using convolutional layers. This is because during convolutions, it is difficult to connect all locations within a feature map \cite{pathak2016context}. This problem is ongoing and numerous solutions have been attempted. For example Pathak et al. \cite{pathak2016context} tried to solve the problem by introducing fully-connected layers to directly connect all activation. Other models \cite{liu2018image,xiong2019foreground,liu2019bidirectional,yu2019free,jam2021r} have used different techniques with Liu et al. \cite{liu2018image} introducing partial convolutions with hard mask updating and Yu et al. \cite{yu2019free}, gated convolutions with soft mask learning \cite{yu2019free}. Though it has shown high understanding of the image structure there are still difficulties in solving problems with masks of arbitrary sizes, particularly in generating facial features with preserved realism of the inpainted regions. Facial features authenticate the face, thus a failure, incorrectly predicted or unnatural removal and replacement of a facial component will be easily detected by the audience (human). In this case, the face with such inpainted regions can be easily classed as invalid or determined as fake.

For these reasons described above, we propose the foreground guided image inpainting network. Our objective is to design and implement a model that has the capability to preserve the prominent features of the face with respect to various expressions and non-natural attributes as seen in Figure \ref{fig:ours}.
To instantiate the design, we assume that the foreground pixels reflect the background ones, which are readily available for disentanglement in latent space and are masked within the input image by the binary mask regions.
The key point to consider from our assumption is that a foreground segmentation masks have a representation of  disentangled pixels in latent space. Thus using a segmented mask manifold \cite{liu2015semantic} will enable the CNN layers to propagate features with respect to facial attributes (natural and non-natural), pose and shapes. 

\section{Related Work}
Pathak et al. \cite{pathak2016context} in 2016, proposed a deep learning method to solve the inpainting problem.
Since its introduction, inpainting results have significantly improved over the years by numerous algorithms which have been documented in literature \cite{elharrouss2019image,jam2020comprehensive}. The advantage of these methods is its capability to utilize latent space given a large data of ground-truth images to learn from and hallucinate missing information from an input (masked image). This kind of learning and semantic understanding is particularly important to inpaint or restore images with natural and complex scenes. Iizuka et al. \cite{iizuka2017globally} improved on this method \cite{pathak2016context} by introducing a local and global discriminator. The use of two discriminators with this approach is to enforce local and global coherence of the inpainted region to the entire image. Based on literature, this method \cite{iizuka2017globally} still uses expensive post-processing (Poisson blending) on the generated image for realistic results. 
\subsection{Foreground Facial Inpainting Framework}
\label{sec:headings}
Semantic scene understanding is an integral part of image inpainting because the hallucination of pixels to recover the damaged regions requires a semantic understanding of the global structure to the target region. The semantic segmentation map of a face can well represent the foreground of the image, where the binary mask is applied to create the damaged region. During hand inpainting, the painter takes into consideration the background pixels and tries to semantically draw a silhouette structure outlining the boundaries before colouring and linking the colour end-nodes or strokes to complete the damaged pixels, thus ensuring consistency with the entire image. Naturally, it is intuitive to consider that an occluded face will normally have two eye spots, a nose and a mouth. Based on this assumption, one can conclude that occlusion reasoning can improve the ability of CNNs to better estimate or hallucinate missing pixels regions created by the binary mask.  

Schulter et al. \cite{schulter2018learning} proposed to use a modified version of PSPNet \cite{zhao2017pyramid} to conduct semantic foreground inpainting task. The network is designed to take two inputs (masked image and mask) based on a two pipeline encoder-decoder network where the reconstructed images are semantics and depth for visible and occluded pixels. 
This method is only limited to inpainting with semantic and depth. 
Lu et al. \cite{lu2020semantic} introduced a max-pooling module and used semantic scene without foreground objects to conduct an inpainting task. The max-pooling module which is designed to fit within the  encoder blocks takes an intermediate feature map, a foreground segmentation mask and binary mask as input to output an inpainted feature map. 
The limitation here is that sharing features between models can improve efficiency but degrade performance.
Lee et al. \cite{lee2020maskgan} created the CelebAMask-HQ segmentation mask dataset as key intermediate representation of facial attributes and proposed a model that can flexibly manipulate these attributes with fidelity preservation. However, because GANs use a discriminator as the examiner, a direct supervision of an occluded region will not be possible if the ground-truth regions behind the binary mask are not available. Based on this assumption, we design our model with a discriminator that will judge the occluded regions to ensure that the inpainted region is realistic and semantically consistent with preserved realism. 

Our proposed method uses foreground segmentation masks within a loss model but not through convolutions. Other models \cite{schulter2018learning,xiong2019foreground,lu2020semantic} apply the binary mask on the foreground masks, and passed through convolutions. 
The following sections describes our proposed facial inpainting framework  and a new loss function that uses foreground mask to ensure fidelity preservation.
\section{Architecture}
\begin{figure*}[!ht]
\centering
\includegraphics[width=12cm,height=8cm,keepaspectratio]{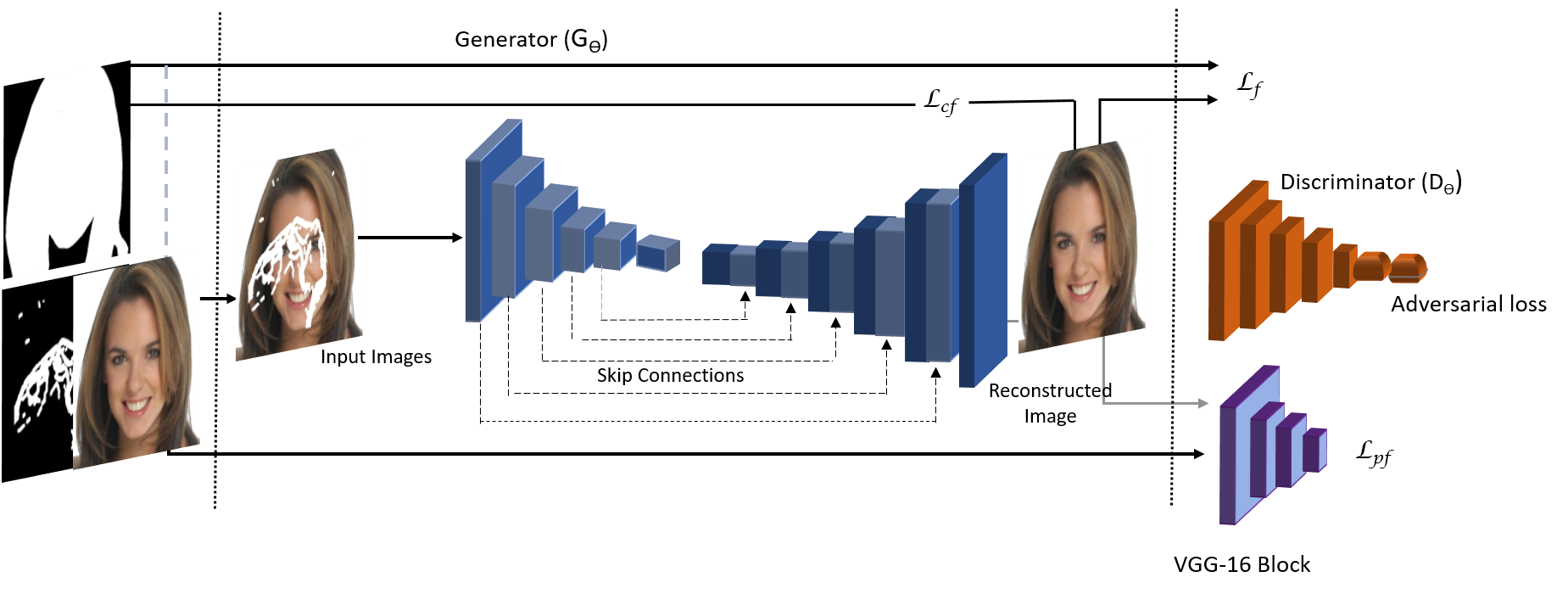}
\caption{\label{fig:foreground_inpaint-framework} Our proposed foreground-guided image inpainting framework with symmetric chain of convolutional and deconvolutional features. The foreground segmentation mask and masked image are the inputs to the network and parameters of the loss functions.}
\end{figure*}
The design of our proposed network has an encoder-decoder, as the generator ($G_{\theta}$) and a discriminator ($D_{\theta}$), to achieve realistic results. The encoder architecture is based on \cite{jam2020symmetric} with the exception of the foreground segmentation mask (henceforth, foreground mask) and masked image as input. During training, we keep the foreground mask in tacked and do not pass it through convolutions. We use the foreground mask within the generator as an access mechanism for the loss to better average pixels during backward pass. The masked image is downsampled to allow the network to learn latent representations of facial feature maps with weak supervision from the foreground mask. We introduce a symmetric chain of convolutional and deconvolutional features, where key components of feature maps without corruption extracted during convolutions are preserved, and added within the reconstruction unit (decoder). 
\subsection{Loss Function} 
\label{sec:loss}
To train our model, we introduce a new loss function based on $L_{2}$  and  $L_{1}$  that takes into consideration  foreground pixels to minimises the error between predicted  $I_{pred}$  and ground-truth ($I_{gt}$) images. 
For a better understanding, we express the loss functions used within the generator as follows:
\begin{equation}
\label{eq:maelossforeground}
\mathcal{L}_{cF} = \frac{1}{N_{I_{gt}}}|| M_{F} \odot (M_{I} -I_{pred})||_1 
\end{equation}
\begin{equation}
\mathcal{L}_{F} = \frac{1}{N_{I_{gt}}}||M_{F} \odot (I_{gt} -I_{pred})||_2
\label{eq:foregroundloss-function}
\end{equation}
where $M_{I}$ is the input, and $N_{I_{gt}}$ is the number of elements in $I_{gt}$ in the shape height ($H$), width ($W$) and channel ($C$), i.e. $N_{I_{gt}} = H*W*C$ and $\odot$ is the element-wise multiplication of the foreground mask $M_{F}$ with $I_{pred}$ and $I_{gt}$. 
These losses ensure preservation of luminance when computing the absolute difference between ground-truth image ($I_{gt}$) and the predicted image ($I_{pred}$). 
\begin{equation}
\mathcal{L}_{p_{F}} = \frac{1}{N_{I_{gt}}}|| M_{F} \odot [\phi_{i}(M_{I}) -\phi_{i}(I_{pred})]||_2
\label{eq:floss-function}
\end{equation}
where $\phi_{i}$ is the feature map of the $i^ith$ layer of pre-trained VGG16 model. The $\mathcal{L}_{p_{F}}$ uses the $M_{F}$ and intermediate features from a fixed VGG16 model  \cite{johnson2016perceptual} to compute the $L_2$ distance between ground-truth and predicted images.

For the discriminator, we adopt the Wasserstein GAN (WGAN) approach to measure the distance between predictions and the ground-truth. 
\begin{equation}
\max_{D}V_{WGAN}=E_{x \sim p_{r}}[(D_{\theta}(I_{gt})]- \\ E_{z\sim p_{z}}[D(G_{\theta}(I_{pred}))] 
\label{eq:fwgandloss}
\end{equation}
Equation \ref{eq:fwgandloss} refers to the WGAN loss based on distributions of $I_{gt}$ (real) data and $I_{pred}$ (generated) data.
\section{Training and Experiments}
\subsection{Datasets}
\begin{figure}[!ht]
\centering
\begin{subfigure}[b]{0.16\linewidth}
	\centering
	\includegraphics[width=\linewidth]{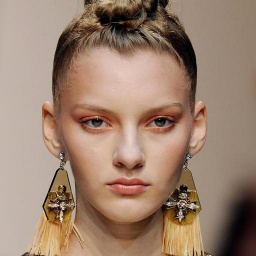}
\end{subfigure}
\begin{subfigure}[b]{0.16\linewidth} 
	\centering
	\includegraphics[width=\linewidth]{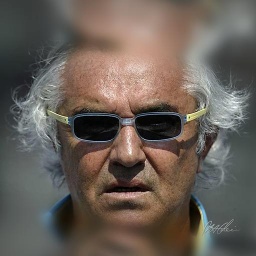}
\end{subfigure}
\begin{subfigure}[b]{0.16\linewidth} 
	\centering
	\includegraphics[width=\linewidth]{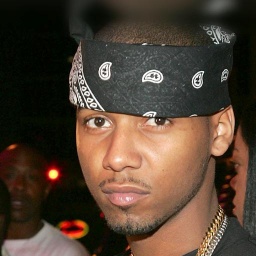}
\end{subfigure}
\begin{subfigure}[b]{0.16\linewidth}
	\centering
	\includegraphics[width=\linewidth]{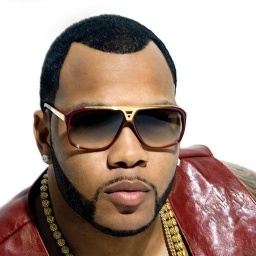}
\end{subfigure}
\begin{subfigure}[b]{0.16\linewidth} 
	\centering
	\includegraphics[width=\linewidth]{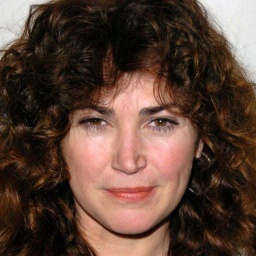}
\end{subfigure}
\begin{subfigure}[b]{0.16\linewidth} 
	\centering
	\includegraphics[width=\linewidth]{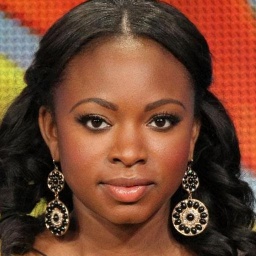}
\end{subfigure}
\\
\begin{subfigure}[b]{0.16\linewidth} 
	\centering
	\includegraphics[width=\linewidth]{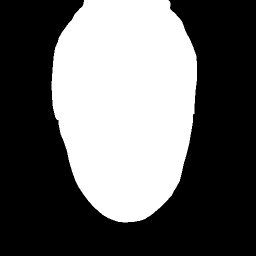}
\end{subfigure}
\begin{subfigure}[b]{0.16\linewidth} 
	\centering
	\includegraphics[width=\linewidth]{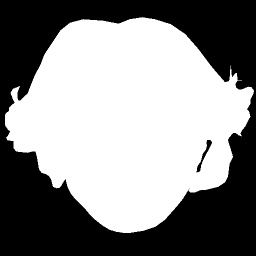}
\end{subfigure}
\begin{subfigure}[b]{0.16\linewidth}
	\centering
	\includegraphics[width=\linewidth]{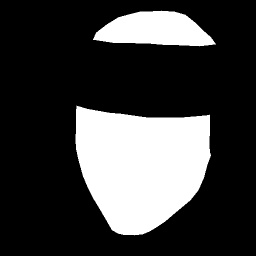}
\end{subfigure}
\begin{subfigure}[b]{0.16\linewidth}
	\centering
	\includegraphics[width=\linewidth]{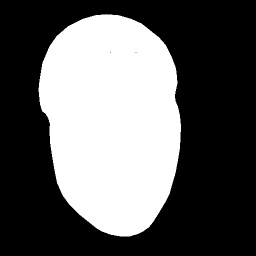}
\end{subfigure}
\begin{subfigure}[b]{0.16\linewidth}
	\centering
	\includegraphics[width=\linewidth]{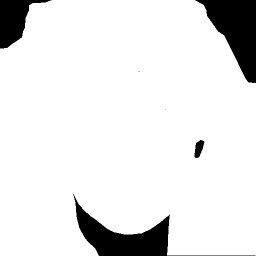}
\end{subfigure}
\begin{subfigure}[b]{0.16\linewidth}
	\centering
	\includegraphics[width=\linewidth]{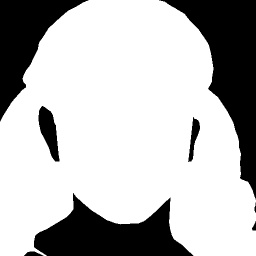}
\end{subfigure}
\caption{\label{fig:fgmask} Images from CelebA-HQ \cite{karras2017progressive} and segmentation masks of face and hair regions from CelebAMask-HQ \cite{lee2020maskgan} used as our foreground masks. The skin region without hair indicates the subject's hair is short or has no hair.}
\end{figure}

The most commonly used publicly available datasets for facial image inpainting is the CelebA-HQ \cite{karras2017progressive}. To create a damaged image or images with missing regions, a binary mask must be applied on the image to simulate the damaged. Usually this is done by an external dataset or function to create these masks regions. We made use of an external dataset namely Quick Draw Mask \cite{iskakov2018semi} resized to $256 \times 256$, with hole-to-image ratios ranging from 0.01 to 0.60. We use semantic segmentation masks from CelebAMask-HQ \cite{lee2020maskgan}. For the case of our model, we extract the skin and hair attributes, to  form the foreground mask shown on Figure~\ref{fig:fgmask} to compute our loss function. 
\subsection{Method Comparison}
Our proposed method compares quantitatively and qualitatively with the state-of-the-art methods.
\begin{itemize}
\item \textbf{Context encoder-decoder framework (CE)} \cite{pathak2016context} introduced the channel-wise fully connected layer .
The channel-wise fully connected layer is designed to directly link all activation; thus enabling propagation of information within the activation of a feature map. 
\item \textbf{Partial Convolution (PConv)} \cite{liu2018image} introduced convolutions with mask updating to alleviate the transfer of feature for irregular masks regions within convolutions. 
\item \textbf{Gated Convolution (GC)} \cite{yu2019free} introduced gating mechanism that learns soft mask within convolutions to ease the transfer of features within convolutions. It is different from PConv in that the irregular mask is learned whereas in former, hard mask is updated in each step.
\item \textbf{Ours} introduces semantic reasoning of features using a foreground mask within the network as a loss model. The key aspect here is that the foreground mask represents disentangled pixels of attribute features of the face. Thus semantic reasoning assists the convolutional layers to hallucinate pixels with fidelity preservation.
\end{itemize}
\subsection{Implementation}
We trained our model with the generator and discriminator loss defined in section~\ref{sec:loss}. We used a similar architecture proposed in \cite{jam2020symmetric} and applied loss weights (coefficients) to the generator loss. Our intention was to ensure that during training, the generator is punished more by increasing its loss weight of the foreground loss to learn structural and textural features to have an overall understanding of the semantic nature of the face region. Our proposed foreground loss further emphasizes the consistency of the predictions by feeding the generator via backward pass with a penalty on the background pixels using the foreground pixels.
We designed our network, and used the Keras libary with Tensorflow-backend to implement and train our model end-to-end. Our choice of dataset made us to follow the experimental settings by the state-of-the-art method \cite{liu2018image} to split our data into 27K train and 3K test images. 
Our model is trained for 100 epochs with a learning rate of $10^{-4}$ in $G(z)$ and $10^{-12}$ in $D(x)$ using the Adam optimizer \cite{kingma2014adam}. Our hardware condition limited us to a batch-size of 5 because of the deep nature of the network. We used GPU support (NVIDIA P6000) to conduct the full experiment from training to inference.

\section{Results and Discussion}
Here we evaluate the performances on predicted image, face/hair regions  and discuss our experiment.

\subsection{Quantitative Results}
In real-life scenarios, the audience appreciate the visual quality of the blending between the inpainted and the original unmasked regions. However, in computer vision, we show quantitative to appreciate different approaches. We use the Mean Square Error (MSE), Mean Absolute Error (MAE), Frenchet Inception Distance (FID), Peak Signal to Noise Ratio (PSNR) and Structure Similarity Index Measure (SSIM), to quantify the performance against the state of the art (\cite{pathak2016context,liu2018image,yu2019free}). Table \ref{tab:table1} shows the quantitative evaluation for the inpainted images with one of ours in bold.
\begin{table}[!h]
\caption{Quantitative comparison of various performance assessment metrics on 3,000 test images from the CelebA-HQ dataset. $\dagger$ Lower is better. $\uplus$ Higher is better.}
\centering
\begin{tabular}{lllllll}
	\toprule
	\multicolumn{2}{c}{Performance Assessment} \\
	\cmidrule(r){1-2}
	Method & Author &MSE $\dagger$& MAE $\dagger$&FID$\dagger$&PSNR $\uplus$&SSIM $\uplus$ \\
	\midrule
	\textbf{CE}& Pathak et al. \cite{pathak2016context}&133.481&129.30&29.96&27.71&0.76\\
	\textbf{PConv}&Liu et al. \cite{liu2018image}& 124.62 &105.94 &15.86 &28.82&0.90 \\
	\textbf{GC}& Yu et al. \cite{yu2019free}&\textbf{102.42} &\textbf{43.10} & \textbf{4.29} &\textbf{39.96}&\textbf{0.92} \\
	\textbf{Ours}& Foreground-guided& 194.86 & 57.38&9.63 & 34.35 &\textbf{0.92}\\
	\bottomrule
\end{tabular}
\label{tab:table1}
\end{table}
The high values obtained for MSE, MAE and FID show poor performance of the model whereas lower values for these metrics indicate better performance. For clarity, we have included on the table $\dagger$ lower is better and $\uplus$ higher is better. PSNR and SSIM with higher values will indicate the prediction is closer to the ground-truth image, which will have a maximum score value of 1 for SSIM.

Our proposed method achieved the best SSIM score (tied with GC) and second performer in the majority of other metrics. This quantitative measures showed our method preserve the structure of the face. To further investigate, we perform quantitative measures on the foreground inpainted face and hair regions only, as shown on Table~\ref{tab:table2}, with our proposed model outperformed the state-of-the-art models.
\begin{table}[!ht]
\caption{Quantitative comparison of various performance assessment metrics on 3,000 test images from the CelebA-HQ dataset on Foreground inpainted regions. $\dagger$ Lower is better. $\uplus$ Higher is better.}
\centering
\begin{tabular}{lllllll}
	\toprule
	\multicolumn{2}{c}{Performance Assessment} \\
	\cmidrule(r){1-2}
	Method & Author &MSE $\dagger$&MAE $\dagger$&FID$\dagger$&PSNR $\uplus$&SSIM $\uplus$ \\
	\midrule
	\textbf{CE}& Pathak et al. \cite{pathak2016context}&133.481&129.30&27.38&27.71&0.76\\
	\textbf{PConv}&Liu et al. \cite{liu2018image}&102.72 &4.35&7.99&29.24&0.87 \\
	\textbf{GC}& Yu et al. \cite{yu2019free}&29.14&\textbf{1.47}&2.23&35.33&0.95\\
	\textbf{Ours}& Foreground-guided& \textbf{26.01} & 2.58&\textbf{1.19} & \textbf{37.38} &\textbf{0.96}\\
	\bottomrule
\end{tabular}
\label{tab:table2}
\end{table}

\subsection{Qualitative Results}
\begin{figure}[!ht]
\centering
\begin{subfigure}[b]{0.16\linewidth}
	\centering
	\includegraphics[width=\linewidth]{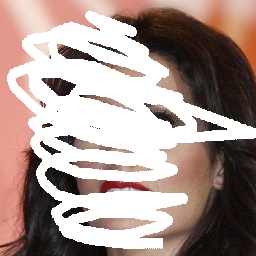}
\end{subfigure}
\begin{subfigure}[b]{0.16\linewidth} 
	\centering
	\includegraphics[width=\linewidth]{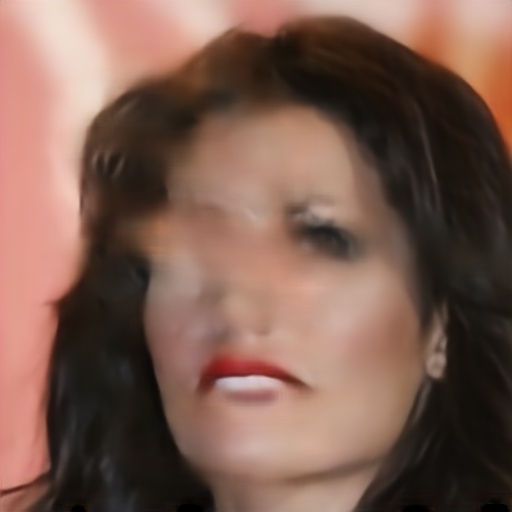}
\end{subfigure}
\begin{subfigure}[b]{0.16\linewidth} 
	\centering
	\includegraphics[width=\linewidth]{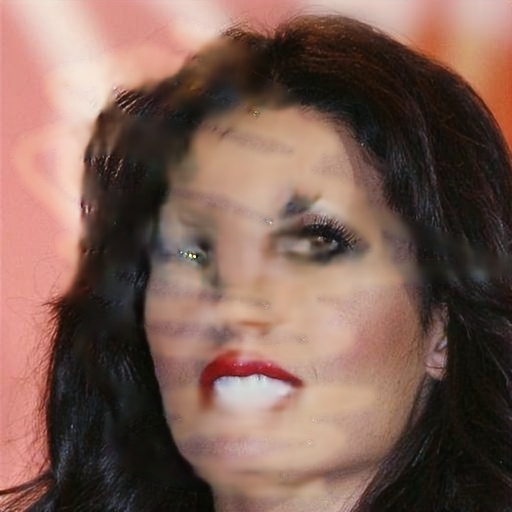}
\end{subfigure}
\begin{subfigure}[b]{0.16\linewidth}
	\centering
	\includegraphics[width=\linewidth]{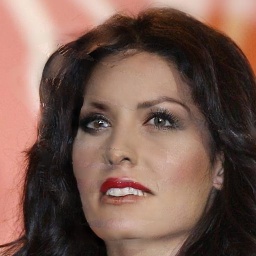}
\end{subfigure}
\begin{subfigure}[b]{0.16\linewidth} 
	\centering
	\includegraphics[width=\linewidth]{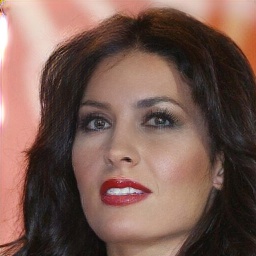}
\end{subfigure}
\begin{subfigure}[b]{0.16\linewidth} 
	\centering
	\includegraphics[width=\linewidth]{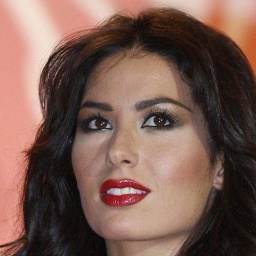}
\end{subfigure}
\\
\begin{subfigure}[b]{0.16\linewidth}        
	\centering
	\includegraphics[width=\linewidth]{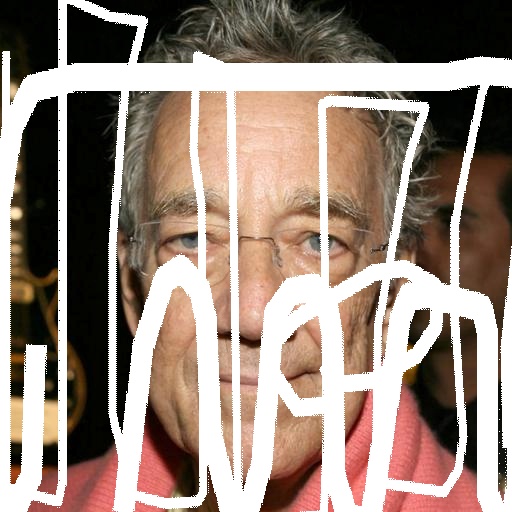}
\end{subfigure}
\begin{subfigure}[b]{0.16\linewidth}        
	\centering
	\includegraphics[width=\linewidth]{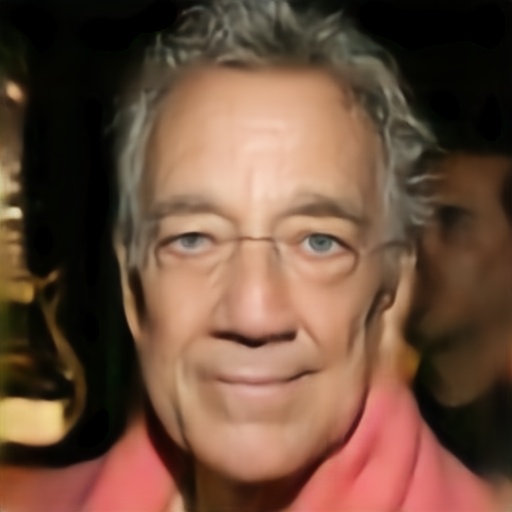}
\end{subfigure}
\begin{subfigure}[b]{0.16\linewidth}        
	\centering
	\includegraphics[width=\linewidth]{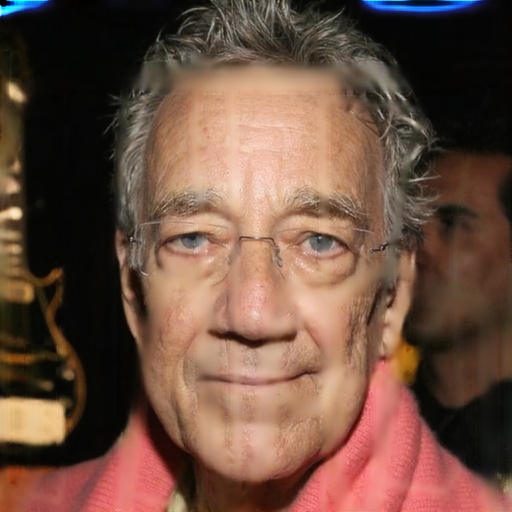}
\end{subfigure}
\begin{subfigure}[b]{0.16\linewidth}        
	\centering
	\includegraphics[width=\linewidth]{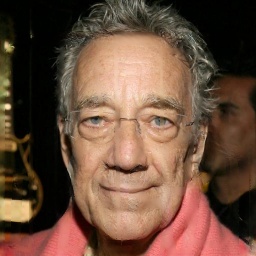}
\end{subfigure}
\begin{subfigure}[b]{0.16\linewidth}        
	\centering
	\includegraphics[width=\linewidth]{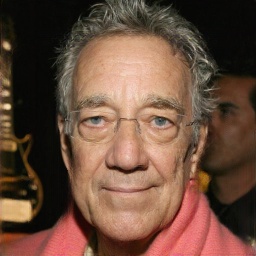}
\end{subfigure}
\begin{subfigure}[b]{0.16\linewidth}        
	\centering
	\includegraphics[width=\linewidth]{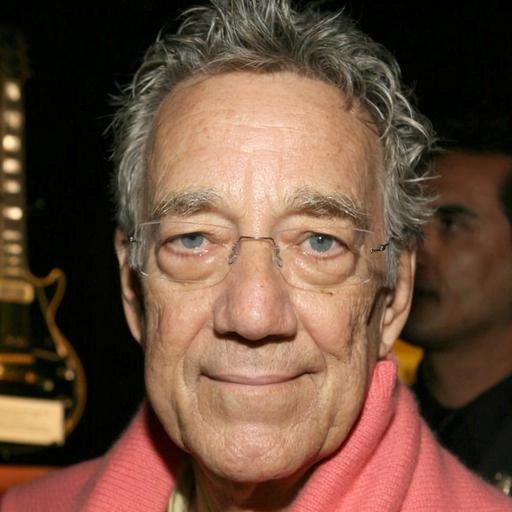}
\end{subfigure}
\\		
\begin{subfigure}[b]{0.16\linewidth} 
	\centering
	\includegraphics[width=\linewidth]{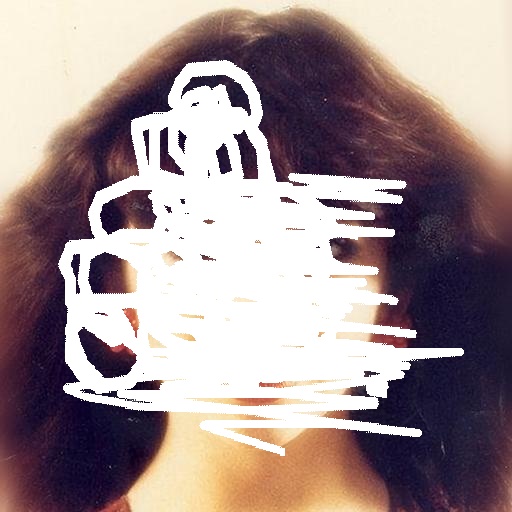}
	\caption{Input}
\end{subfigure}
\begin{subfigure}[b]{0.16\linewidth} 
	\centering
	\includegraphics[width=\linewidth]{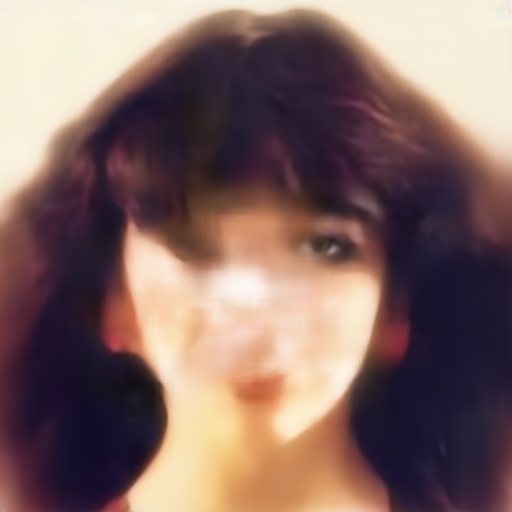}
	\caption{CE}
\end{subfigure}
\begin{subfigure}[b]{0.16\linewidth}
	\centering
	\includegraphics[width=\linewidth]{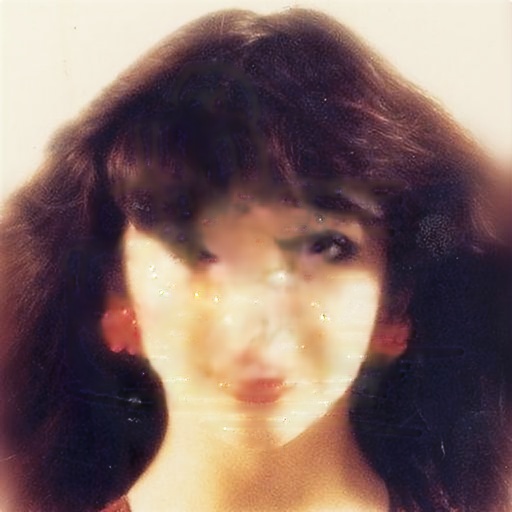}
	\caption{PConv}
\end{subfigure}
\begin{subfigure}[b]{0.16\linewidth}
	\centering
	\includegraphics[width=\linewidth]{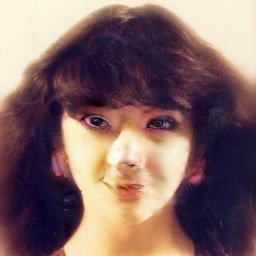}
	\caption{GC}
\end{subfigure}
\begin{subfigure}[b]{0.16\linewidth}
	\centering
	\includegraphics[width=\linewidth]{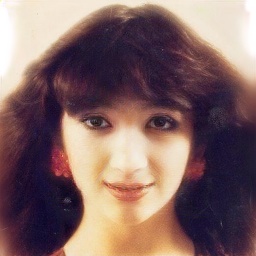}
	\caption{Ours}
\end{subfigure}
\begin{subfigure}[b]{0.16\linewidth}
	\centering
	\includegraphics[width=\linewidth]{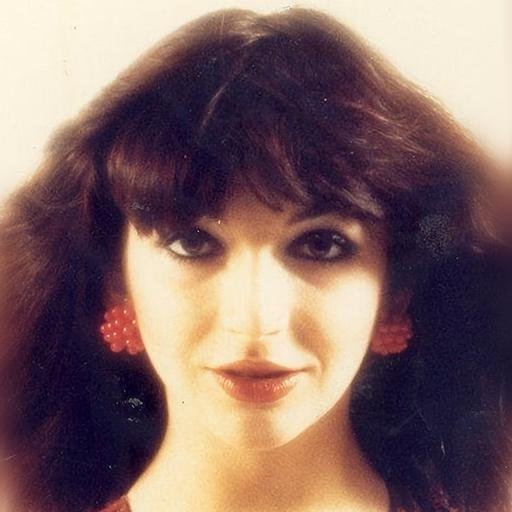}
	\caption{GT}
\end{subfigure}
\caption{\label{fig:resultsfailure}Qualitative comparison of our proposed method with the state-of-the-art methods: (a) \textbf{Input masked-image}; (b) \textbf{CE} \cite{pathak2016context}; (c) \textbf{PConv} \cite{liu2018image}; (d) \textbf{GConv}; (e) \textbf{Ours}; and (f) Ground-truth image.}
\end{figure}

In this section, we show visual comparison with our method compared with the state-of-the-art. Without bias and based on code availability, we used Pathak et al. \cite{pathak2016context} (\textbf{CE}), Liu et al. \cite{liu2018image} (\textbf{PConv}), Yu et al. \cite{yu2019free} (\textbf{GC}) to measure against our model. From Figure~\ref{fig:resultsfailure}, \textbf{CE} struggles with arbitrary hole-to-image mask regions and the generated image is blurry, while \textbf{PConv} and \textbf{GC} leave a bit of artefacts (best viewed when zoomed) on the generated image. 

\subsection{Semantic Inpainting with Fidelity Preservation}
\begin{figure}[!ht]
\centering
\begin{subfigure}[b]{0.16\linewidth} 
	\centering
	\includegraphics[width=\linewidth]{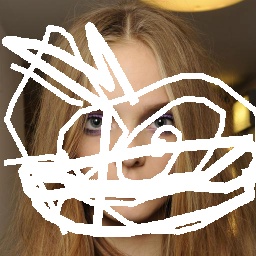}
	\caption{Input}
\end{subfigure}
\begin{subfigure}[b]{0.16\linewidth} 
	\centering
	\includegraphics[width=\linewidth]{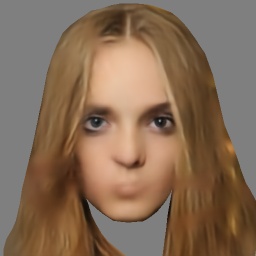}
	\caption{CE}
\end{subfigure}
\begin{subfigure}[b]{0.16\linewidth}
	\centering
	\includegraphics[width=\linewidth]{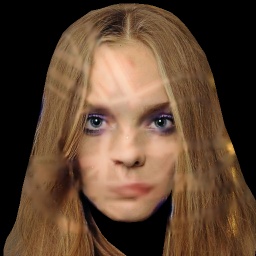}
	\caption{PConv}
\end{subfigure}
\begin{subfigure}[b]{0.16\linewidth}
	\centering
	\includegraphics[width=\linewidth]{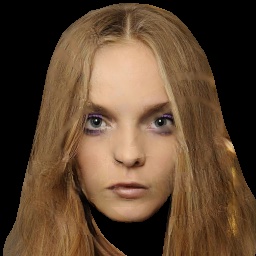}
	\caption{GC}
\end{subfigure}
\begin{subfigure}[b]{0.16\linewidth}
	\centering
	\includegraphics[width=\linewidth]{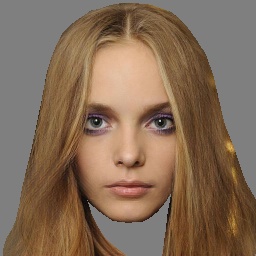}
	\caption{Ours}
\end{subfigure}
\begin{subfigure}[b]{0.16\linewidth}
	\centering
	\includegraphics[width=\linewidth]{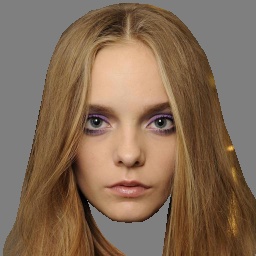}
	\caption{GT}
\end{subfigure}
\caption{\label{fig:fgcompare}Qualitative comparison of foreground inpainted regions with the state-of-the-art methods: (a) \textbf{Input masked-image}; (b) \textbf{CE} \cite{pathak2016context}; (c) \textbf{PConv} \cite{liu2018image}; (d) \textbf{GConv}; (e) \textbf{Ours}; and (f) Ground-truth image.}
\end{figure}

The qualitative results in Figure~\ref{fig:resultsfailure} showed the performance of our model has great visual quality when compared to the state of the art. To further show reasonable semantic understanding of predictions, we showed that our model can fill-in high-level textural and structural information as seen in Figure~\ref{fig:resultsfailure} where other methods have failed. As seen on the Figure~\ref{fig:ours}, the lip region on the image inpainted by our model is fully recovered, indicating a full semantic understanding of the image by putting a broader smile as compared to the original input. Furthermore, on Figure~\ref{fig:resultsfailure}, the earring, nose and eye regions are fully recovered with preserved realism with our model. This also show that our model has high semantic understanding of facial features when trained with the joint loss function. Thus semantic understanding of features in latent space significantly improves the visual quality of generated facial components, which is further supported by Figure \ref{fig:fgcompare}. When focuses on the face and hair regions only, the first row of Figure \ref{fig:fgcompare} illustrates the ability of our method in predicting the missing eye region and the others show accurate prediction of mouth regions.

\section{Conclusion }
In this paper, we introduced a method to inpaint missing region(s) within an image using foreground guidance. The results we obtained suggests the importance of foreground guidance training for the prediction of challenging corrupted patches on an image. We have shown how our model is able to predict and reconstruct plausible and realistic features with preserved realism of the faces. Our model can produce high visual quality results that meets the objective of real-world in the wild scenarios. Further exploitation of foreground pixels is a promising ground-work for future inpainting task.
\bibliographystyle{unsrt}
\bibliography{references}  

%
%
%
%

\end{document}